\documentclass[3p,preprint,12pt]{elsarticle}

\usepackage{amsmath}
\usepackage{amssymb}
\usepackage{hyperref}
\usepackage{enumitem}
\hypersetup{
    colorlinks=true,
    linkcolor=blue,
    citecolor=blue,
    urlcolor=blue,
    pdftitle={Agentic Root Cause Analysis through Evidence-Grounded Reasoning without Fault Samples},
    pdfauthor={Amaury Wei and Olga Fink},
}

\journal{Reliability Engineering \& System Safety}

\begin{document}

\begin{frontmatter} \vspace*{-2cm}

\title{Agentic Root Cause Analysis through Evidence-Grounded Reasoning without Fault Samples}

\author[imos]{Amaury Wei\fnref{orcid-aw}}
\author[imos]{Olga Fink\corref{cor1}\fnref{orcid-of}}
\cortext[cor1]{Corresponding author: \texttt{olga.fink@epfl.ch}}
\fntext[orcid-aw]{ORCID: \href{https://orcid.org/0000-0002-8626-9128}{0000-0002-8626-9128}}
\fntext[orcid-of]{ORCID: \href{https://orcid.org/0000-0002-9546-1488}{0000-0002-9546-1488}}

\affiliation[imos]{organization={EPFL},
            addressline={IMOS Laboratory},
            city={Lausanne},
            postcode={CH-1015},
            country={Switzerland}}

\begin{abstract}
Diagnosing the root cause of anomalies is essential for safe and reliable industrial operation. Despite extensive sensor instrumentation, identifying abnormal behavior and determining its underlying physical root cause remain challenging. Existing data-driven approaches aim to automate this process, but often provide limited justification for their diagnoses and rely on labeled examples of faulty operation that are scarce in industrial settings. To address this gap, we propose AgentRCA, an agentic framework for evidence-grounded root cause analysis. AgentRCA operates without labeled examples of faulty operation, using normal-operation data and natural-language descriptions of candidate fault mechanisms. Rather than learning fault-specific mappings, AgentRCA performs inference-time reasoning by combining a data-driven digital twin (modeling normal system behavior) with a tool-augmented large language model. The agent evaluates competing physical fault hypotheses using quantitative diagnostic evidence and engineering knowledge, producing ranked diagnoses with traceable, natural-language justifications. Evaluations on a real-world multiphase-flow facility and a large-scale chemical plant demonstrate competitive diagnostic performance when labeled fault data are scarce. Ablation studies highlight the importance of quantitative evidence and structured engineering knowledge. These results establish AgentRCA as a framework for diagnosing industrial faults from normal-operation data while making the evidence underlying each diagnosis explicit and inspectable. Code available at \url{https://github.com/AmauryWEI/agentrca}.
\end{abstract}



\begin{keyword}
Root cause analysis\sep Fault diagnosis\sep Large language models\sep Agentic AI
\end{keyword}

\end{frontmatter}

\section{Introduction}

Modern industrial systems are equipped with dense sensor networks and control systems that continuously monitor the state of complex assets~\cite{qin2012survey,isermann2005fault}. These measurements have enabled significant advances in system monitoring and health management~\cite{hu2022prognostics}, allowing abnormal operating conditions to be detected from high-dimensional process data~\cite{pang2021deep,zhao2025data}. However, detecting that a system behaves abnormally is only the first step toward effective maintenance. To determine the appropriate intervention, engineers need to identify the root cause underlying the observed abnormal behavior~\cite{sarker2021data}.\vspace{3pt}

\newpage

\vspace*{-2.35\baselineskip}
This process of identifying the underlying physical fault mechanism is known as root cause analysis (RCA). In practice, engineers approach this task iteratively: they formulate candidate fault hypotheses, examine relevant evidence, and progressively refine or rule out competing explanations~\cite{isermann2005fault,venkatasubramanian2003review_part1}. RCA requires integrating statistical evidence from measurements with engineering knowledge of operating conditions, system topology, control interactions, and fault propagation. Each step is guided by the current hypotheses, which determine what evidence to examine next. Individual symptoms are often insufficient to discriminate between plausible causes, making diagnosis dependent on the consistency of multiple pieces of evidence. This task is particularly challenging in closed-loop industrial systems, where different fault mechanisms can produce similar symptoms and the same fault can manifest differently depending on operating conditions and controller actions~\cite{chiang2012fault,yin2014review}.\vspace{3pt}

Automating this diagnostic process is particularly challenging because the information available in industrial settings is inherently asymmetric. Normal operation is routinely observed and extensively recorded, whereas representative fault examples are scarce, particularly for failures that are rare, safety-critical, or costly to reproduce~\cite{ramirez2023semi,deng2024semi,gao2024semi}. Yet industrial systems remain rich in engineering knowledge: engineers understand component functions, sensor semantics, system topology, control interactions, and the basic physical consequences of known failure mechanisms. This knowledge, however, is not naturally incorporated by most data-driven diagnostic methods.\vspace{3pt}

Data-driven methods automate fault diagnosis by learning diagnostic relationships directly from historical process measurements, typically with limited explicit use of the engineering knowledge used by human experts~\cite{zhao2025data}. Supervised approaches~\cite{4700287,wang2023causal,zhao2024deep} typically learn direct mappings from measurements to predefined fault classes, achieving strong diagnostic performance when extensive labeled fault data are available. Unsupervised approaches~\cite{sakurada2014anomaly,jang2021adversarial,hu2025deep} remove this dependence on fault labels by learning normal system behavior, while more recent methods primarily identify statistical deviations or their likely sources~\cite{budhathoki2022causal,nagalapatti2025robust,floreale2024sensitivity,chen2025causal}. However, jointly exploiting normal-operation data and engineering knowledge for root cause analysis remains comparatively underexplored.\vspace{3pt}

Recent advances in large language models (LLMs) provide new opportunities to integrate engineering expertise with data-driven analysis. LLMs have demonstrated promising capabilities for reasoning over scientific knowledge across diverse domains~\cite{chen2026bridging,lupoiu2025multi,jin2025agentmd}, while tool-augmented agents can extend this reasoning to quantitative evidence by selecting external tools, interpreting their outputs, and using the resulting information to guide subsequent analyses~\cite{yao2023react,schick2023toolformer}. These capabilities have motivated growing interest in LLM-based approaches for maintenance and fault diagnosis~\cite{lai2025fr}. However, many existing approaches retain a data-driven paradigm similar to supervised fault diagnosis: they rely on historical fault examples or retrieved incident records~\cite{ahmed2023recommending,chen2024automatic,wang2024rcagent}, task-specific fine-tuning on fault data~\cite{wang2025diagllm,lin2025fd,zheng2024empirical}, or continued human input during diagnosis~\cite{zhang2025llm}. Consequently, they do not fundamentally address the scarcity of labeled fault data that limits supervised diagnosis. In contrast, approaches that avoid fault data~\cite{KHAN2025109152} provide the LLM only with the variables flagged by a PCA-based fault detector, and misdiagnose faults whose causes are not reflected in these variables. Therefore, the use of LLM agents to diagnose faults by combining engineering knowledge with quantitative evidence, without relying on historical fault examples, remains largely unexplored.\vspace{3pt}

To address this gap, we introduce AgentRCA, an agentic framework for evidence-grounded root cause analysis that combines a data-driven digital twin with a tool-augmented LLM. The digital twin models normal system behavior and is learned exclusively from normal-operation data, while the LLM is provided with the type of engineering knowledge routinely available to maintenance engineers, including system overview, sensor positioning, and qualitative descriptions of candidate fault mechanisms. Rather than learning fault-specific mappings from labeled examples, AgentRCA performs inference-time reasoning over this engineering knowledge and measured evidence. The agent iteratively gathers statistical evidence through diagnostic tools, compares competing fault hypotheses, and rules out explanations that are inconsistent with the observed system behavior. In this way, AgentRCA reflects the iterative diagnostic workflow of maintenance engineers, producing diagnostic traces that explicitly connect the hypotheses considered to the evidence examined. AgentRCA therefore operates in a \emph{zero-shot fault diagnosis} setting, without labeled examples of faulty operation.\vspace{3pt}

We evaluate AgentRCA on two closed-loop industrial systems spanning different scales, dynamics, and fault mechanisms: the real-world PRONTO multiphase-flow facility~\cite{pronto} and the Tennessee Eastman chemical process~\cite{tep}. Without access to labeled fault data, AgentRCA achieves strong diagnostic performance and remains competitive with supervised approaches when only limited fault data are available. We further examine how diagnostic performance is affected by the agent’s reasoning strategy, the evidence available through its diagnostic tools, and the engineering knowledge provided to the LLM. Together, these results demonstrate the potential of combining normal-operation data with available engineering knowledge to support maintenance engineers in diagnosing the root cause of industrial faults when labeled fault observations are scarce or unavailable.\vspace{3pt}

Our main contributions can be summarized as:\vspace{-6pt}
\begin{itemize}[noitemsep,leftmargin=*]
    \item We propose AgentRCA, a zero-shot root cause analysis framework in which a tool-augmented LLM agent reasons over engineering knowledge expressed in natural language and over quantitative evidence from a data-driven digital twin learned exclusively from normal-operation data.
    \item We design a set of complementary diagnostic tools that characterize reconstruction errors, statistical shifts, changes in variable relationships, and temporal signatures, using operating-condition matching and regime-specific references of normal operation.
    \item We design a diagnosis-oriented prompt structure that integrates engineering knowledge with quantitative evidence to evaluate competing fault hypotheses and produce inspectable diagnostic traces.
    \item We evaluate AgentRCA on a real multiphase-flow facility and the Tennessee Eastman Process, and compare it with an LLM baseline without tool augmentation and with supervised classifiers across varying amounts of labeled fault data, quantifying how much labeled fault data supervised methods require to match zero-shot diagnosis performance.
    \item Through ablations of the reasoning protocol, the engineering knowledge provided, the LLM, and auxiliary technical documentation, we identify which components drive diagnostic performance.
\end{itemize}

\section{Related work}

\subsection{Model- and knowledge-based fault diagnosis}

Model- and knowledge-based methods constitute traditional approaches to fault diagnosis. Model-based methods compare observed system behavior with predictions from first-principles~\cite{venkatasubramanian2003review_part1} or physics-based models~\cite{patton1989fault,gao2015real} to detect and isolate faults. Knowledge-based approaches instead encode relationships between faults and observable symptoms through expert rules~\cite{venkatasubramanian2003review_part2,zhang1994frequency}, fault trees~\cite{vesely1981fault,bartlett2009integrated}, Bayesian networks~\cite{weber2012overview}, or other qualitative models~\cite{venkatasubramanian2003review_part2}. While interpretable, these traditional approaches require accurate physical models or manually constructed knowledge representations that can become complex to maintain and adapt to changing operating conditions or new fault mechanisms.

Recent hybrid approaches combine engineering knowledge with operational data, for example by converting pFMEA knowledge into a predefined failure-mode tree~\cite{mokhtarzadeh2025framework} or by fusing manually engineered domain-knowledge features with features learned from sensor data~\cite{su2024knowledge}. AgentRCA instead uses natural-language fault knowledge at inference time to guide the iterative collection and evaluation of diagnostic evidence, without encoding it into a fixed rule or probabilistic structure.

\subsection{Data-driven fault diagnosis, anomaly detection, and root cause analysis}

Supervised fault diagnosis is commonly formulated as a time-series classification problem, where models learn to discriminate predefined fault classes from historical sensor measurements. Recent approaches use deep architectures to capture temporal dependencies~\cite{zhao2024deep} and graph-based models to represent interactions between process variables~\cite{wang2023causal,liu2024causal}. However, these methods require representative labeled examples of the faults to be diagnosed. Semi-supervised methods~\cite{ramirez2023semi,deng2024semi} and small-sample approaches~\cite{gao2024semi} reduce this requirement but still rely on fault-specific training examples.

Unsupervised methods instead learn normal system behavior using, for example, autoencoder-based representations~\cite{sakurada2014anomaly,jang2021adversarial,hu2025deep} or conditional generative models~\cite{chen2025anomaly}, and detect deviations without labeled fault data. Beyond anomaly detection, recent methods localize faults or identify likely root causes through sensitivity analysis~\cite{floreale2024sensitivity}, statistical attribution~\cite{budhathoki2022causal,nagalapatti2025robust}, or causal dependencies between process variables~\cite{han2025root,chen2025causal}. However, localizing the source of an anomaly does not necessarily identify the underlying physical fault mechanism, as different faults can produce deviations in the same process variables. AgentRCA instead uses these statistical deviations as evidence for evaluating competing physical fault hypotheses.

\subsection{Large language models for root cause analysis}

Most LLM-based approaches to RCA rely on prior fault data. Some retrieve and reason over historical incidents~\cite{ahmed2023recommending,chen2024automatic,wang2024rcagent}, while others fine-tune the LLM on fault-specific data~\cite{wang2025diagllm,lin2025fd,zheng2024empirical}, or combine signal-to-text representations with fault examples~\cite{lai2025fr}. Other frameworks require continued interaction with human experts during diagnosis~\cite{zhang2025llm}. Moreover, retrieval-based approaches have primarily been developed for software and cloud systems, where diagnosis relies on textual logs and incident reports rather than on continuous sensor measurements from closed-loop physical processes~\cite{10301257}. Their dependence on historical fault information limits their applicability to industrial systems, where such data are scarce or unavailable.

Approaches without historical fault data must instead extract diagnostic evidence from available measurements. One approach is to provide raw time series directly to LLMs, typically serialized as text. Although LLMs can forecast numerical sequences zero-shot~\cite{gruver2023llmtime} and detect certain anomalies~\cite{alnegheimish2024sigllm,zhou2025llmanomalies}, identifying  the causes of observed time-series behavior remains challenging~\cite{merrill2024struggle}, and their advantages over specialized time-series models are not yet firmly established~\cite{tan2024useful}. FaultExplainer~\cite{KHAN2025109152} follows this strategy by using an LLM  to interpret PCA-derived variable contributions without requiring historical fault data. However, PCA contributions primarily identify variables associated with abnormal behavior, which may not provide enough evidence to distinguish between competing fault mechanisms. FaultExplainer also assumes that a fault has already been detected.

AgentRCA addresses these limitations by combining natural-language engineering knowledge with quantitative evidence, without requiring historical fault data. Rather than interpreting raw time series or relying on a single diagnostic method, it investigates a set of complementary tools that extract evidence against regime-specific references of normal operation, allowing the LLM to evaluate competing fault hypotheses. Including normal operation among the candidate states also enables joint fault detection and diagnosis.

\section{Method}

AgentRCA combines a normal-operation digital twin, quantitative diagnostic tools, and an LLM agent equipped with engineering knowledge to evaluate candidate physical fault mechanisms (Fig.~\ref{fig:method_overview}). The digital twin provides a reference for normal system behavior, while the diagnostic tools extract quantitative evidence from sensor measurements. The agent then uses this evidence to iteratively evaluate fault hypotheses and identify the mechanism that best explains the observed behavior.

\subsection{AgentRCA framework}

Let $\mathbf{x}\in\mathbb{R}^{T\times M}$ denote a multivariate observation
window spanning $T$ time steps and $M$ process variables. We consider a
closed set of candidate system states
$\mathcal{C}=\{c_0,c_1,\ldots,c_K\}$, where $c_0$ denotes normal operation
and $c_1,\ldots,c_K$ correspond to candidate physical fault mechanisms.
Each candidate is associated with a natural-language description,
forming the set $\mathcal{D}=\{d_0,d_1,\ldots,d_K\}$.
The diagnostic objective is to rank the candidates in $\mathcal{C}$
according to their consistency with the observed system behavior.\vspace{3pt}

\begin{figure}[t]
    \centering
    \vspace{-0.8cm}
    \includegraphics[width=1.0\linewidth]{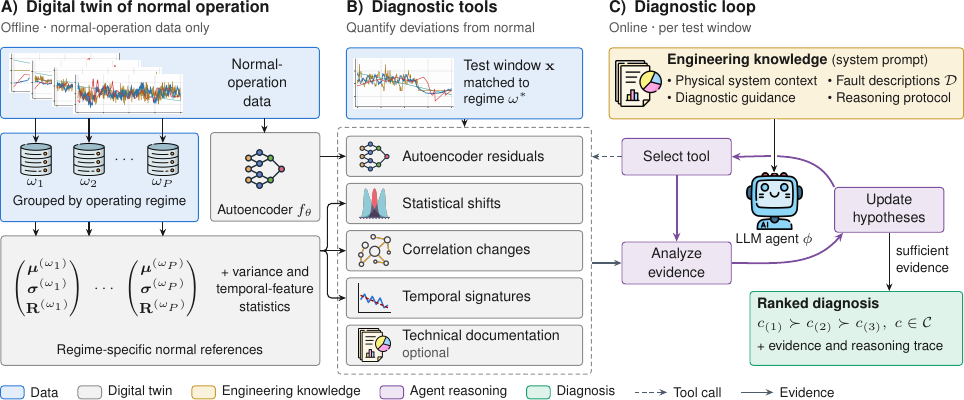}
    \vspace{-0.6cm}
    \caption[Architecture overview of AgentRCA]{\textbf{Architecture overview of AgentRCA.} \textbf{(A)} Offline, a data-driven digital twin is learned from normal-operation data only. Normal windows are grouped into operating regimes $\omega_1, \dots, \omega_P$ defined by system setpoints, for which reference statistics are precomputed (means $\boldsymbol{\mu}$, standard deviations $\boldsymbol{\sigma}$, correlation matrices $\mathbf{R}$, \dots). A global autoencoder $f_\theta$ is trained across all regimes. \textbf{(B)} Diagnostic tools quantify how a test window $\mathbf{x}$ deviates from the normal reference of its matched regime $\omega^{*}$; technical documentation can optionally replace the structured engineering descriptions in the agent's prompt. \textbf{(C)} Online, the LLM agent $\phi$ receives engineering knowledge, including the fault descriptions $\mathcal{D}$. It iteratively selects tools, analyzes the returned evidence and updates its hypotheses, then returns a ranked diagnosis over the candidate states $\mathcal{C}$ with supporting evidence.}
    \label{fig:method_overview}
    \vspace{-0.1cm}
\end{figure}

AgentRCA performs this diagnosis using a tool-augmented LLM agent $\phi$ interacting with a data-driven digital twin of normal operation (Fig.~\ref{fig:method_overview}). The digital twin combines operating-regime-specific statistical references with a learned autoencoder model of normal multivariate behavior. Both components are constructed exclusively from normal-operation measurements recorded from the monitored system.\vspace{3pt}

During diagnosis, the agent receives engineering knowledge about the monitored system in natural language: a description of the plant, its sensors and control variables, and the qualitative physical signatures expected for each candidate fault mechanism (Section~\ref{subsec:knowledge}). For example, a leakage fault may cause a downstream pressure decrease and corresponding changes in connected flow measurements. 

This knowledge reflects information typically available in system specifications, operating manuals, and failure mode, effects, and criticality analyses (FMECAs), and is routinely used by maintenance engineers to formulate fault hypotheses and identify relevant evidence. In our experiments, the prompts were derived from official system and process descriptions, with information that could reveal diagnostic labels removed. They provide qualitative knowledge of fault mechanisms, not labeled faulty-operation measurements or empirical fault signatures learned from historical examples.

\subsection{Data-driven digital twin and diagnostic tools}

The data-driven digital twin provides a reference for how the monitored system behaves during normal operation. Because process variables may exhibit substantially different distributions under different operating conditions, a single global normal reference is insufficient (demonstrated in Section~\ref{subsec:results_faithfulness}). We therefore partition the normal-operation data into operating regimes
$\Omega=\{\omega_1,\omega_2,\ldots,\omega_P\}$ defined from system setpoints. For example, operating regimes may be defined by turbine power setpoints in a hydropower plant or combinations of feed-rate and temperature setpoints in a chemical processing plant.

For each regime $\omega\in\Omega$, we compute the expected mean vector
$\boldsymbol{\mu}^{(\omega)}\in\mathbb{R}^M$, standard deviation vector
$\boldsymbol{\sigma}^{(\omega)}\in\mathbb{R}^M$, and pairwise Pearson correlation matrix
$\mathbf{R}^{(\omega)}\in\mathbb{R}^{M\times M}$. In parallel, an autoencoder $f_\theta$ is trained on normal-operation windows from all regimes to learn shared multivariate temporal patterns. At inference time, the observed window is matched to the normal operating regime $\omega^\ast$ whose setpoint vector has the smallest standardized Euclidean distance to that of the observed window. The tools described below then quantify deviations relative to either this regime-specific reference or the learned normal manifold.

The diagnostic tools are selected to provide complementary views of departures from normal operation: multivariate anomaly evidence from a learned normal-behavior model, together with interpretable statistical, correlation, and temporal descriptors that characterize how the observed behavior differs from its normal reference. This separation also allows alternative normal-behavior models, such as PCA, to replace the autoencoder without changing the agent’s diagnostic workflow, as demonstrated in the Supplementary Information.

\begin{figure}[t]
    \centering
    \includegraphics[width=1.0\linewidth]{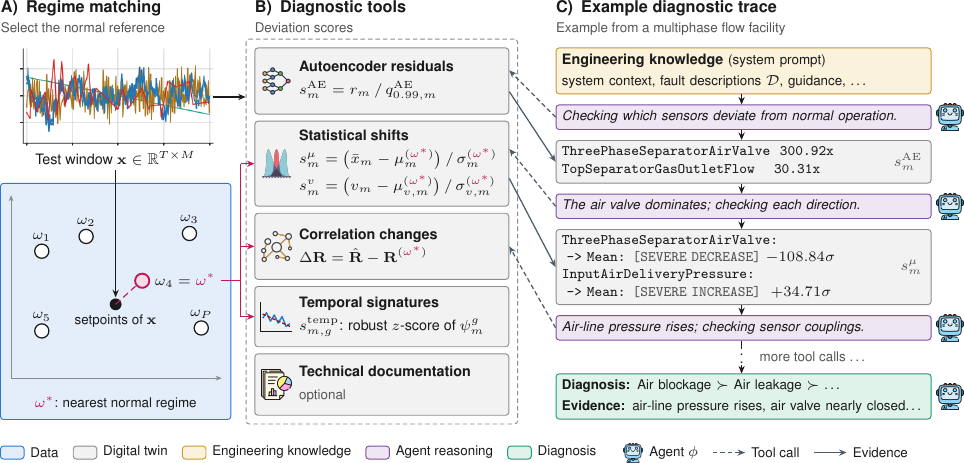}
    \caption[Diagnostic tools and example diagnostic trace]{\textbf{Diagnostic tools and example diagnostic trace.} \textbf{(A)} The test window $\mathbf{x}$ is matched to the normal operating regime $\omega^{*}$ with the nearest setpoints in standardized distance. \textbf{(B)} Each diagnostic tool computes deviation scores of $\mathbf{x}$ relative to normal operation (Eqs.~\eqref{eq:ae}--\eqref{eq:temporal_sign}). Statistical shifts, correlation changes and temporal signatures are computed against the references of $\omega^{*}$, whereas autoencoder residuals are normalized by thresholds of the global model $f_\theta$. \textbf{(C)} Shortened excerpt of a diagnostic trace on the multiphase flow facility (PRONTO). Tool outputs are reproduced verbatim (top entries only) and the agent's reasoning is paraphrased. Guided by the engineering knowledge in its system prompt, the agent $\phi$ queries the tools one at a time, interprets the returned scores, and returns a ranked diagnosis with supporting evidence.}
    \label{fig:method_details}
\end{figure}

\paragraph{Autoencoder residuals}
The autoencoder tool detects deviations from expected normal operation by identifying multivariate temporal patterns that are poorly reconstructed by a model trained on normal data. Unlike tools based on individual statistical descriptors, the autoencoder can identify deviations from learned multivariate temporal patterns, including nonlinear relationships that may not be reflected in individual mean or variance shifts. The model $f_\theta$ is a one-dimensional convolutional encoder--decoder operating on standardized windows represented as $M\times T$ arrays. For each process variable $m$, its reconstruction residual and normalized anomaly score are:
\begin{equation} \label{eq:ae}
r_m = \frac{1}{T}\sum_{t=1}^{T}(z_{t,m}-\hat{z}_{t,m})^2,
\qquad
q^{\mathrm{AE}}_{0.99,m} = Q_{0.99}\left(\{r_m^{(n)}\}_{n=1}^{N}\right),
\qquad
s^{\mathrm{AE}}_m = \frac{r_m}{q^{\mathrm{AE}}_{0.99,m}+\epsilon}.
\end{equation}

Here, $z_{t,m}$ and $\hat{z}_{t,m}$ denote the standardized observation and its reconstruction, respectively; $Q_{0.99}$ is the empirical 99th percentile of reconstruction residuals over normal training windows; and $\epsilon=10^{-7}$ ensures numerical stability. Thus, $s^{\mathrm{AE}}_m>1$ approximately indicates that the reconstruction error exceeds the variable-specific normal reference. The tool returns the ten variables with the largest normalized residuals.

\paragraph{Statistical shifts}
While autoencoder residuals indicate which variables deviate from learned normal behavior, they do not directly reveal the direction or nature of these deviations. The statistical-shift tool complements them by identifying increases or decreases in the mean and variability of individual process variables relative to the matched operating regime. For each variable $m$, the window mean and its signed deviation from normal operation are:

\begin{equation} \label{eq:mean_shift}
\bar{x}_m = \frac{1}{T}\sum_{t=1}^{T}x_{t,m},
\qquad
s^\mu_m = \frac{\bar{x}_m-\mu_m^{(\omega^\ast)}}{\sigma_m^{(\omega^\ast)}}.
\end{equation}

Positive and negative values of $s^\mu_m$ indicate increases and decreases relative to normal operation, respectively. Variability changes are assessed by comparing the within-window variance with the distribution of window variances under the matched normal regime:
\begin{equation}\label{eq:variance_shift}
v_m = \frac{1}{T}\sum_{t=1}^{T}(x_{t,m}-\bar{x}_m)^2,
\qquad
s^v_m = \frac{v_m-\mu_{v,m}^{(\omega^\ast)}}{\sigma_{v,m}^{(\omega^\ast)}},
\end{equation}

where $\mu_{v,m}^{(\omega)}$ and $\sigma_{v,m}^{(\omega)}$ are the mean and standard deviation of the within-window variances computed from normal training windows belonging to regime $\omega$. Positive and negative variance-shift scores indicate increased and decreased variability, respectively. The tool ranks variables by $\max(|s^\mu_m|,|s^v_m|)$ and returns the ten largest deviations, allowing the agent to distinguish, for example, pressure loss from pressure increase or increased fluctuations from steady offsets.

\paragraph{Correlation changes}
Individual-variable deviations do not fully describe how a fault affects the relationships between interconnected process variables. The correlation tool therefore identifies changes in their statistical dependencies, which may reveal inconsistencies in the expected joint behavior of the system even when individual measurements remain close to their normal ranges. For example, if airflow increases without the corresponding pressure response normally observed in the system, this relationship change may provide evidence consistent with an air leak. The tool computes the sample correlation matrix $\hat{\mathbf{R}}$ of the observed window and compares it with the regime-specific normal matrix:
\begin{equation}\label{eq:correlation_shift}
\Delta\mathbf{R} = \hat{\mathbf{R}}-\mathbf{R}^{(\omega^\ast)}.
\end{equation}

Each discrepancy $|\Delta R_{ij}|$ is normalized by the 99th percentile of the absolute correlation shifts observed for the same variable pair over normal windows of the matched regime, and the tool returns the ten variable pairs with the largest normalized discrepancies. These changes provide additional evidence for evaluating candidate fault mechanisms given the known physical relationships between variables.

\paragraph{Temporal signatures}
While the preceding tools primarily summarize deviations over the observation window, the temporal-signature tool examines how individual signals evolve within it. It identifies transient excursions, gradual trends, and changes in short-term dynamics that may be obscured by window-level statistics. For example, a progressive pressure decrease may suggest a developing leak, whereas repeated pressure fluctuations may indicate intermittent flow disturbances. For each queried variable $m$, the tool computes the within-window range, linear slope, and lag-one autocorrelation, normalized against their distributions over normal training windows from the matched operating regime $\omega^*$:

\begin{equation} \label{eq:temporal_sign}
\begin{aligned}
\psi^{\mathrm{range}}_m
&= \max_t x_{t,m}-\min_t x_{t,m},
&\qquad
\psi^{\mathrm{slope}}_m
&= \frac{\sum_{t=1}^{T}(t-\bar{t})(x_{t,m}-\bar{x}_m)}
{\sum_{t=1}^{T}(t-\bar{t})^2},\\[4pt]
\psi^{\mathrm{lag1}}_m
&= \operatorname{corr}(x_{1:T-1,m},x_{2:T,m}),
&\qquad
s^{\mathrm{temp}}_{m,g}
&= \frac{\psi^g_m-\operatorname{median}_n(\psi^{g,(n)}_m)}
{\operatorname{IQR}_n(\psi^{g,(n)}_m)}.
\end{aligned}
\end{equation}

Here, $g\in\{\mathrm{range},\mathrm{slope},\mathrm{lag1}\}$ denotes the temporal feature, and $\operatorname{IQR}$ is the interquartile range of that feature across normal training windows from the operating regime $\omega^*$.\vspace{3pt}

These tools provide complementary quantitative evidence for assessing candidate fault mechanisms; they do not assign fault labels. During inference, the agent selects tools based on to its current hypotheses, with a suggested order provided in the system prompt (Fig.~\ref{fig:prompt_structure}). Each diagnosis is limited to ten tool-call iterations, and all experimental runs completed within this limit. Additional evidence sources, such as technical documentation, can be incorporated through the same interface without retraining the LLM or modifying the digital twin.

\subsection{Engineering knowledge and agent reasoning} \label{subsec:knowledge}

\begin{figure}[t]
    \centering
    \vspace{-0.4cm}
    \includegraphics[width=1.0\linewidth]{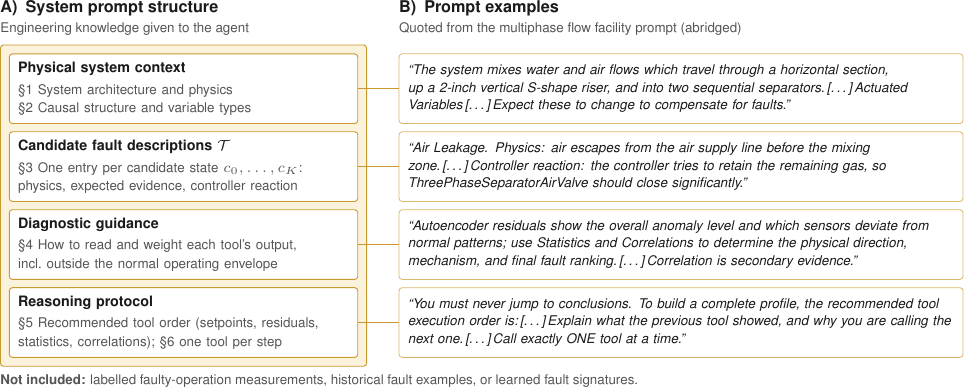}
    \vspace{-0.8cm}
    \caption[Engineering knowledge in the system prompt]{\textbf{Engineering knowledge in the system prompt.} \textbf{(A)} The prompt has four functional components: the physical system context (architecture, causal structure and variable types), a description of each candidate state $c_0,\dots,c_K$ forming $\mathcal{D}$ (physical mechanism, expected process behavior and controller reaction), guidance on how to interpret and weight each tool's output, and a reasoning protocol with a recommended tool execution order. \textbf{(B)} Abridged verbatim excerpts of each component from the multiphase flow facility (PRONTO) prompt. The prompt contains only qualitative engineering knowledge: no labelled faulty-operation measurements, historical fault examples or learned fault signatures.}
    \label{fig:prompt_structure}
\end{figure}

AgentRCA organizes the engineering knowledge and diagnostic instructions provided to the LLM into four components (Fig.~\ref{fig:prompt_structure}): physical system context, descriptions of candidate fault mechanisms, guidance for interpreting diagnostic evidence, and a reasoning protocol.

Variable names in the system context are kept consistent with the diagnostic-tool interfaces, allowing the agent to relate engineering knowledge directly to measured evidence. Each fault description $d_k$ specifies the physical mechanism, its expected effects on process variables and, where relevant, the anticipated controller response. These descriptions define diagnostic hypotheses without prescribing a fixed mapping from observed deviations to fault labels. The diagnostic guidance explains how to interpret the different evidence sources, for example by treating correlation changes as supporting rather than primary evidence.\vspace{-4pt}

\paragraph{Hypothesis-tracking protocol}
During inference, the agent maintains a ranked table of candidate fault mechanisms with supporting and contradicting evidence, updating it as diagnostic tools are queried. This protocol structures the evaluation of competing hypotheses and provides a traceable record of the evidence supporting the final diagnosis.\vspace{3pt}

The complete prompts used in the experiments are provided in the Supplementary Material. The ablation experiments additionally consider an unstructured protocol, a ReAct-like protocol~\cite{yao2023react}, a prompt without tool-usage guidance, and a configuration with additional domain-informed diagnostic rules intended to prevent common misdiagnoses. These variants isolate the effects of reasoning structure and diagnostic guidance; the default configuration uses hypothesis tracking with the zero-shot prompt.


\subsection{Technical documentation as a source of engineering knowledge} \label{subsec:method_docs}

To assess the importance of diagnosis-specific prompt design, we investigate whether AgentRCA can instead reason directly over technical documentation not structured for diagnosis. Such documents often contain unevenly distributed information, including details irrelevant to fault diagnosis. In this variant, the prompt retains only the candidate state names, recommended tool order, and execution rules, while the complete technical document replaces the engineered system and fault descriptions. Each technical document is supplied to the agent in full, without further rewriting or summarization, apart from removing ground-truth fault labels. The diagnostic tools remain unchanged, and the same documentation is provided to the LLM baseline without tools (Section~\ref{subsec:setup_baselines}).

\section{Experimental setup} \label{sec:setup}

\subsection{Benchmark systems and fault scenarios}

We evaluate AgentRCA on two complementary closed-loop process-monitoring benchmarks: the PRONTO multiphase-flow facility~\cite{pronto} and the Tennessee Eastman Process (TEP)~\cite{tep}. The benchmarks differ substantially in scale and complexity, allowing us to evaluate the framework on both real experimental measurements and a larger simulated chemical process.\vspace{3pt}

\begin{figure}[t]
    \centering
    \vspace{-0.5cm}
    \includegraphics[width=1.0\linewidth]{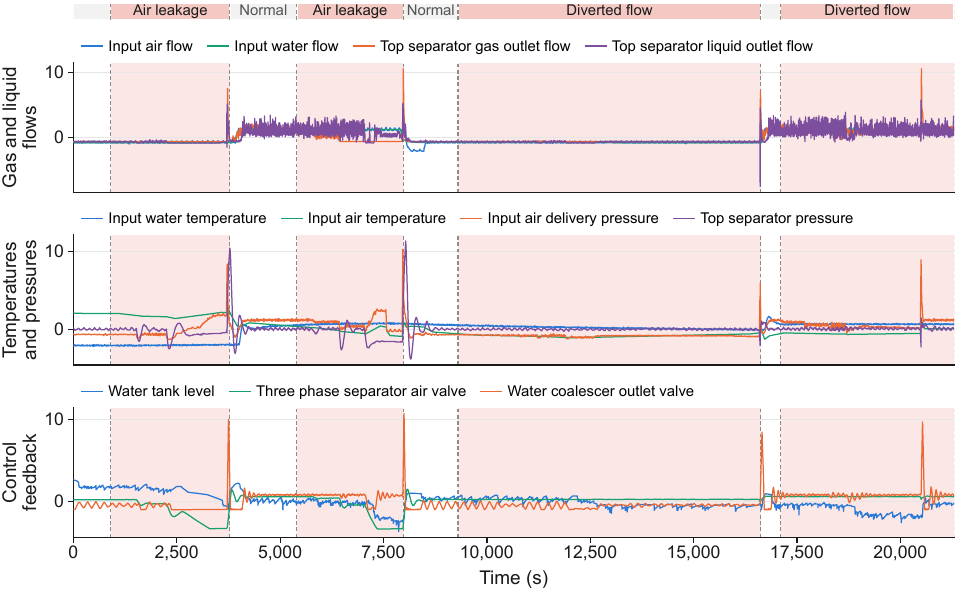}
    \vspace{-0.8cm}
    \caption[Example recording from the PRONTO dataset]{\textbf{Example recording from ``Testday3'' of the PRONTO dataset~\cite{pronto}.} The strip above the panels shows the annotated operating state over time; fault intervals are shaded light red in the signal panels. Signals are a subset of the flow, temperature, pressure, level and valve measurements, grouped by type and standardized to Z-scores, so they should be read by their relative dynamics rather than their physical units.}
    \label{fig:pronto_overview}
\end{figure}

PRONTO contains measurements from a physical multiphase-flow facility, which records 17 process variables. All diagnostic methods use 11 variables describing the gas–liquid flow path from the inlets to the separators. The dataset provides annotated intervals of normal operation and four induced fault mechanisms: air blockage, air leakage, diverted flow, and slugging. The faults are introduced with increasing severity, resulting in fault signatures that evolve over time rather than forming only abrupt transitions. An example is shown on Fig.~\ref{fig:pronto_overview}.

\begin{figure}[t]
    \centering
    \vspace{-0.8cm}
    \includegraphics[width=1.0\linewidth]{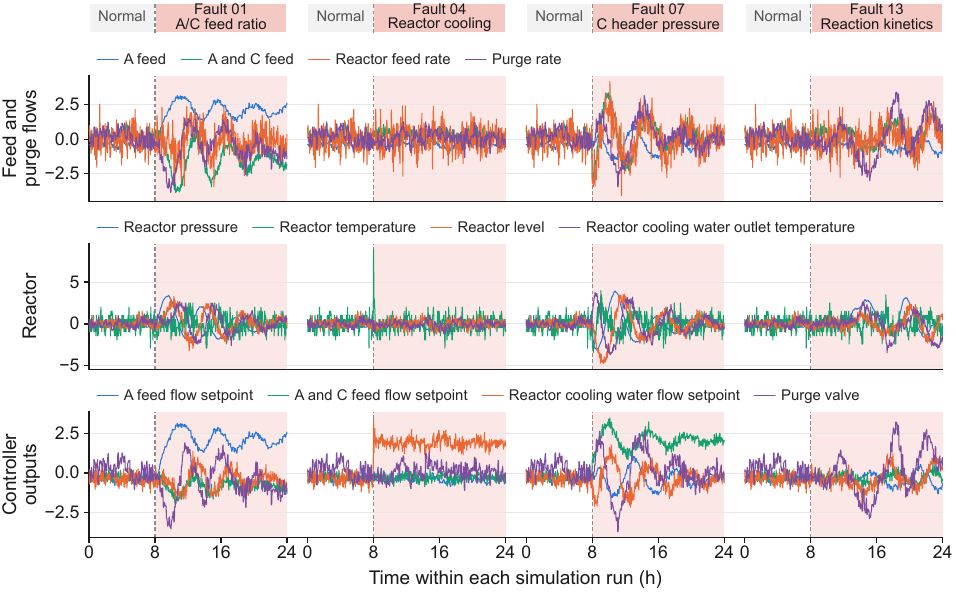}
    \vspace{-0.8cm}
    \caption[Example simulation runs from the TEP dataset]{\textbf{Example simulation runs from the Tennessee Eastman Process (TEP) dataset~\cite{tep}.} Four test runs, each with a single fault introduced after 8~h of normal operation, are shown one after another (first 24~h of each run). The strip above the panels shows the operating state; fault intervals are shaded light red. Signals are standardized to Z-scores and should be read by their relative dynamics rather than their physical units.}
    \label{fig:tep_overview}
    \vspace{-0.1cm}
\end{figure}

TEP is a simulated closed-loop chemical process containing 52 measured and manipulated variables and 20 predefined fault scenarios, numbered according to the original TEP reference~\cite{tep}. Fault examples are shown on Fig.~\ref{fig:tep_overview}. We evaluate normal operation together with 12 faults (Faults 1, 2, 4--8, and 10--14), yielding 13 candidate states. Faults 16--20 are excluded because their physical causes are unspecified. Faults 3, 9, and 15 are excluded from the primary evaluation because previous studies~\cite{YIN20121567} and our supplementary analysis indicate weak or inconsistent diagnostic signatures under the adopted evaluation protocol.\vspace{3pt}

For both benchmarks, the candidate diagnostic set is defined before evaluation and includes normal operation. AgentRCA therefore performs closed-set diagnosis: each test window is ranked against the supplied set of candidate physical fault mechanisms. Fault labels from the evaluation set are used only to score the returned diagnoses; they are not provided to the agent as training examples or retrieved during inference.

\subsection{Data preprocessing and evaluation protocol}

Each multivariate recording is segmented into non-overlapping fixed-length windows and assigned the annotated operating condition. We use windows of 60 samples (60\,s) for PRONTO~\cite{pronto} and 20 samples (1\,h) for TEP~\cite{tep}.

PRONTO does not provide an official train--test split. We therefore construct a balanced test set by sampling 21 windows from each of the five evaluated operating conditions (normal operation and four faults). The remaining windows define the training pool. Importantly, AgentRCA uses only the normal-operation subset of this training pool to construct its digital twin. Faulty training windows are used only by the supervised comparison methods. For TEP, we use the benchmark's official training--testing split and again restrict AgentRCA's digital-twin training data to normal-operation windows.

All normalization statistics (autoencoder input scaling and regime-specific references) are estimated from normal-operation training windows only and applied unchanged to test windows. Window counts per condition and split are given in the Supplementary Material.

Throughout the paper, \emph{zero-shot diagnosis} refers specifically to the absence of labeled faulty-operation measurements for training, model adaptation, or retrieval. AgentRCA does observe the test window being diagnosed and receives natural-language descriptions of the candidate fault mechanisms and system context. The term therefore does not imply diagnosis without prior engineering knowledge.

\subsection{Comparison methods} \label{subsec:setup_baselines}

\paragraph{LLM baseline without tools} We evaluate a direct zero-shot LLM baseline that receives the dataset's technical documentation and the raw multivariate window serialized as text~\cite{gruver2023llmtime,alnegheimish2024sigllm}. The LLM returns a single diagnosis among the same candidate states without access to diagnostic tools. This comparison assesses direct LLM interpretation of raw measurements against AgentRCA’s evidence-grounded approach. The technical-documentation variant of AgentRCA (Section~\ref{subsec:method_docs}) additionally enables comparison under the same engineering knowledge, isolating the contribution of tool-derived evidence relative to raw measurements.\vspace{-6pt}

\paragraph{Supervised baselines} To contextualize the performance obtained without labeled fault examples, we compare AgentRCA with three established supervised approaches representing complementary time-series classification paradigms: LightGBM~\cite{ke2017lightgbm}, an autoencoder classifier, and MiniRocket~\cite{dempster2021minirocket} with Ridge classification~\cite{hoerl1970ridge}. These comparisons are not intended to place the methods under identical supervision: the supervised models are explicitly given labeled examples of the evaluated faults, whereas AgentRCA is not. Trained on all available labeled fault data, they provide a practical upper performance reference attainable by data-driven methods on these benchmarks under optimal supervision conditions. Trained on progressively smaller subsets, they allow us to quantify how much labeled fault data is needed to match AgentRCA’s zero-shot performance.\vspace{-6pt}

\paragraph{MiniRocket} Each window is transformed into 10{,}000 features~\cite{dempster2021minirocket}, which are classified with a cross-validated Ridge classifier with balanced class weights.\vspace{-6pt}

\paragraph{Autoencoder classifier} The autoencoder classifier uses the same encoder architecture as AgentRCA’s digital twin, with an added classification head. It is trained end-to-end using class-weighted cross-entropy combined with a reconstruction loss: $\mathcal{L}=\mathrm{CE}_{w}(\mathrm{logits},y)+0.25\mathrm{MSE}(\hat{x},x)$. Its stopping epoch is selected using the evaluation set, so its scores serve as an optimistic performance reference.\vspace{-6pt}

\paragraph{LightGBM} Each window is represented by 18 statistical and temporal features per process variable (e.g., quantiles, linear trend, step changes and lag-one autocorrelation) and classified with a class-weighted gradient-boosted tree ensemble.\vspace{4pt}

All hyperparameters are listed in the Supplementary Material. For the sample-efficiency experiments, the supervised methods are trained on progressively smaller labeled training sets. On PRONTO, an equal number of windows is drawn per class, capped by its own available windows. On TEP, windows are sampled without replacement to obtain each training-set size, with stratification by class: all fault classes are equally represented, while normal operation, which has more training data, accounts for approximately 13.5\% of each training subset. Every class receives at least one window when the training-set size is at least the number of classes.

\subsection{Implementation}

Unless otherwise stated, the agent $\phi$ is instantiated with DeepSeek-V4.1-Flash~\cite{xu2026deepseek} at a sampling temperature of 1.0 and high reasoning effort, with at most ten tool calls per diagnosis. We chose this model because its weights are openly available, keeping the experiments reproducible and allowing the agent to run on local infrastructure without transmitting confidential plant data. The model ablation (Section~\ref{subsec:model_robustness}) evaluates other LLMs, whose setup is described in the Supplementary Material, and shows that AgentRCA works across different LLM backbones, although diagnostic accuracy varies between models. Operating regimes are defined by unique combinations of the air and water flow setpoints for PRONTO, and by the chemical feed setpoints for TEP.

For the technical-documentation experiments, we use the report distributed with\linebreak PRONTO~\cite{pronto}, removing ground-truth fault labels while leaving the remaining content unchanged. As TEP has no comparable report, we use GPT-5.6-Sol to compile a technical document from the original process description~\cite{tep}, covering the process overview, variable definitions and units, operating constraints, and disturbances with their affected quantities and forms, but no diagnostic signatures.

\subsection{Evaluation metrics}

Diagnostic performance is evaluated using Top-1 and Top-2 accuracy. Top-1 accuracy measures the proportion of held-out windows for which the highest-ranked candidate matches the annotated operating condition. Top-2 accuracy measures the proportion for which the annotated condition appears among the two highest-ranked candidates. Because both evaluation sets contain the same number of windows per class, Top-1 accuracy equals balanced accuracy. Top-2 accuracy complements Top-1 accuracy by indicating whether the correct mechanism remains among the two highest-ranked hypotheses, even when it is not the top prediction.

All agentic methods (AgentRCA and the LLM baseline) are evaluated over three independent runs, while supervised baselines are evaluated over five random seeds. Reported uncertainties, including error bars and shaded regions, indicate one standard deviation across runs under fixed data splits.

\section{Results}

We first evaluate AgentRCA against an LLM baseline and supervised classifiers on the PRONTO (Section~\ref{subsec:results_pronto}) and TEP case studies (Section~\ref{subsec:results_tep}), including the amount of labeled fault data supervised methods require to reach comparable accuracy. We then analyze the diagnostic evidence supporting AgentRCA's root cause analysis (Section~\ref{subsec:results_faithfulness}), the effects of LLM choice and reasoning protocol (Section~\ref{subsec:results_ablations}), and the impact of replacing diagnosis-specific prompts with technical documentation (Section~\ref{subsec:results_knowledge}). Unless stated otherwise, AgentRCA uses DeepSeek-V4.1-Flash with the hypothesis protocol.

\subsection{Zero-shot diagnosis approaches supervised performance on PRONTO} \label{subsec:results_pronto}

\begin{figure}[t]
    \centering
    \includegraphics[width=1.0\linewidth]{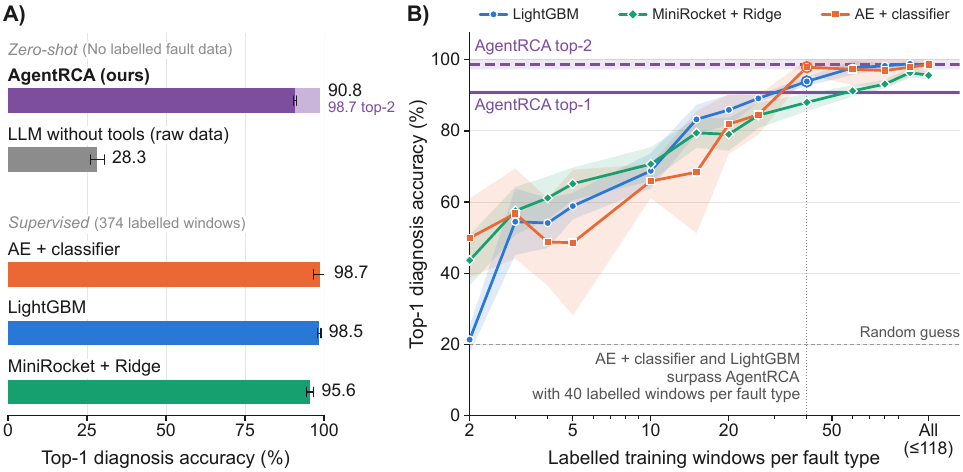}
    \vspace{-0.8cm}
    \caption[Root cause analysis on the PRONTO dataset]{\textbf{Root cause analysis on the PRONTO dataset~\cite{pronto}.} \textbf{(A)} Top-1 diagnosis accuracy of zero-shot methods (AgentRCA and the LLM without tools) and of supervised baselines trained on all labeled training windows; the light part of the AgentRCA bar indicates its Top-2 accuracy. Error bars indicate one standard deviation across runs. \textbf{(B)} Top-1 accuracy of the supervised baselines as a function of the number of labeled training windows per fault type (log scale), with shaded bands indicating one standard deviation across runs. Classes with fewer training windows contribute all of them; the full training set contains between 26 and 118 windows per class. Horizontal lines show AgentRCA's zero-shot Top-1 and Top-2 accuracy.}
    \label{fig:pronto_accuracy}
\end{figure}

On PRONTO, AgentRCA achieves $90.8 \pm 0.6\%$ Top-1 and $98.7 \pm 1.5\%$ Top-2 accuracy across the five operating states (Fig.~\ref{fig:pronto_accuracy}A), using only normal-operation data and textual fault descriptions. Supervised baselines trained on the full labeled training set reach 98.7\% (autoencoder classifier), 98.5\% (LightGBM), and 95.6\% (MiniRocket). AgentRCA therefore approaches fully supervised performance without labeled fault examples, with Top-1 accuracy within 4.8–7.9 percentage points of the supervised baselines, ranking the correct state among its top two hypotheses in nearly all cases. In contrast, the LLM baseline, which receives technical documentation and raw measurements but no diagnostic tools, achieves only $28.3 \pm 2.2\%$ Top-1 accuracy, barely above random guessing (20\%). These results demonstrate the value of combining structured engineering knowledge with digital-twin diagnostic tools for accurate zero-shot diagnosis.

We then quantified how much labeled data the supervised baselines need to reach this level (Fig.~\ref{fig:pronto_accuracy}B). Their accuracy decreases substantially when labeled data become scarce: with five labeled windows per class, the best baseline reaches 65.1\%. LightGBM and the autoencoder classifier first exceed AgentRCA's Top-1 accuracy with 40 labeled windows per class (182 windows, about 3\,h of labeled operation, of which 2.4\,h are faulty operation). Only LightGBM matches AgentRCA's Top-2 accuracy, with 100 windows per class (354 windows, about 6\,h). In practice, obtaining labeled examples of each fault mechanism may require costly fault-induction experiments or documented fault occurrences,  limiting the practical deployment of supervised methods when such data are scarce. AgentRCA instead produces ranked diagnoses supported by inspectable quantitative evidence, without requiring historical fault examples.

\begin{figure}[t]
    \centering
    \vspace{-0.4cm}
    \includegraphics[width=1.0\linewidth]{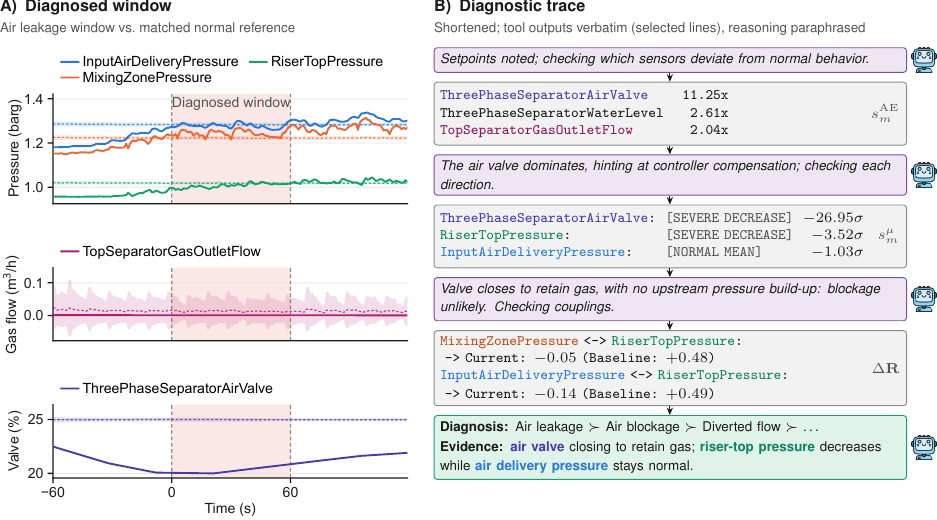}
    \vspace{-0.9cm}
    \caption[Example diagnosis of an air leakage on PRONTO]{\textbf{Example diagnosis of an air leakage on PRONTO.} \textbf{(A)} Diagnosed 60\,s window (shaded, between dashed lines) with one window of context on each side. Solid lines show the measured signals; dashed lines and bands show the mean $\pm$1 standard deviation of normal training windows recorded at the same air and water flow setpoints. \textbf{(B)}~Shortened diagnostic trace produced by AgentRCA for this window. Tool outputs are reproduced verbatim (selected lines) and the agent's reasoning is paraphrased; sensor names are colored as in (A). The autoencoder flags the three-phase separator air valve as the dominant deviation. The statistical-shift tool shows that the valve closes sharply while the riser-top pressure decreases and the air delivery pressure remains normal, which argues against an air blockage. Correlation changes provide additional evidence of altered relationships between the air supply and the riser. The agent ranks air leakage first, which matches the annotated state.}
    \label{fig:pronto_trace}
\end{figure}

Fig.~\ref{fig:pronto_trace} illustrates how complementary evidence helps distinguish competing fault mechanisms. The autoencoder identifies the three-phase separator air valve as the dominant deviation, while statistical shifts and correlation changes reveal a pattern consistent with air leakage rather than air blockage. In particular, the absence of an upstream pressure increase argues against blockage despite the pronounced valve response. The trace makes the quantitative evidence supporting this distinction directly inspectable.

\subsection{AgentRCA scales to a plant-wide chemical process} \label{subsec:results_tep}

\begin{figure}[t]
    \centering
    \vspace{-0.8cm}
    \includegraphics[width=1.0\linewidth]{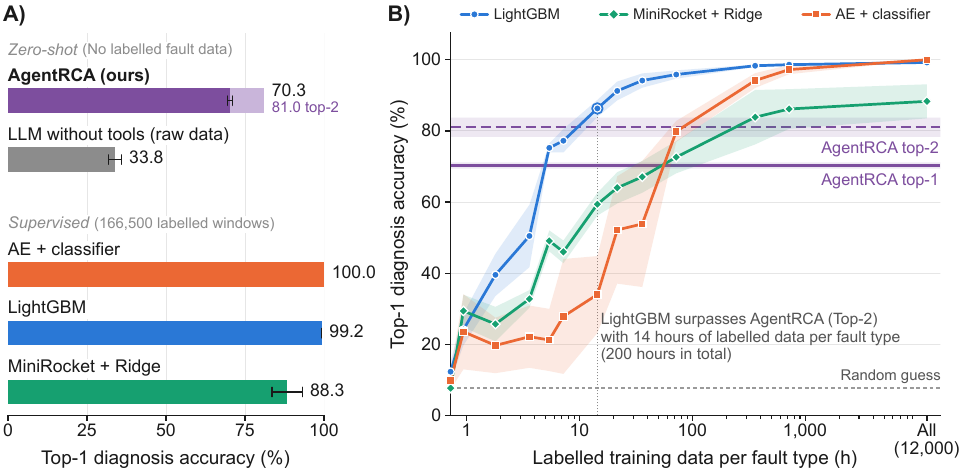}
    \vspace{-0.8cm}
    \caption[Root cause analysis on the TEP dataset]{\textbf{Root cause analysis on the TEP benchmark~\cite{tep}.} \textbf{(A)} Top-1 diagnosis accuracy of zero-shot methods (AgentRCA and the LLM without tools, which receives the raw window) and of supervised baselines trained on all labeled training windows; the light part of the AgentRCA bar indicates its Top-2 accuracy. Error bars indicate one standard deviation across runs. \textbf{(B)} Top-1 accuracy of the supervised baselines as a function of the hours of labeled training data per fault type (log scale; one window covers 1\,h), with shaded bands indicating one standard deviation across runs. Training subsets preserve the class proportions of the training set, in which normal operation is represented about twice as often as each fault. Horizontal lines show AgentRCA's zero-shot Top-1 and Top-2 accuracy.}
    \label{fig:tep_accuracy}
    \vspace{-0.2cm}
\end{figure}

TEP presents a more challenging setting, with 52 coupled variables, 13 candidate states, and fault effects that propagate through recycle streams and control loops. Despite this complexity, AgentRCA achieves $70.3 \pm 0.9\%$ Top-1 and $81.0 \pm 2.7\%$ Top-2 accuracy without labeled fault examples, substantially above the 7.7\% random-guess rate (Fig.~\ref{fig:tep_accuracy}A). The LLM baseline without tools reaches $33.8 \pm 2.0\%$ Top-1 accuracy, while fully supervised classifiers reach 100.0\% (autoencoder classifier), 99.2\% (LightGBM), and 88.3\% (MiniRocket). Although AgentRCA remains below fully supervised performance, it discriminates among 13 candidate operating states using only normal-operation references and engineering knowledge, while providing quantitative evidence for its diagnoses.

The advantage of zero-shot diagnosis becomes clearer when labeled fault data are limited (Fig.~\ref{fig:tep_accuracy}B). With approximately 2\,h of labeled data per fault type (25 windows in total), the best supervised baseline achieves only 39.5\%. LightGBM, the most sample-efficient baseline, first exceeds AgentRCA's Top-1 accuracy with about 5\,h of labeled data per fault type (75\,h in total), but requires about 14\,h per fault type (200\,h in total) to exceed AgentRCA's Top-2 reference. The autoencoder classifier and MiniRocket require about 72\,h per fault type (1{,}000\,h in total) to exceed AgentRCA's Top-1 accuracy. These results highlight the value of AgentRCA when labeled fault data are scarce: the correct mechanism frequently remains among the leading hypotheses, which are accompanied by quantitative evidence that can guide further investigation.

\begin{figure}[t]
    \centering
    \vspace{-1.1cm}
    \includegraphics[width=1.0\linewidth]{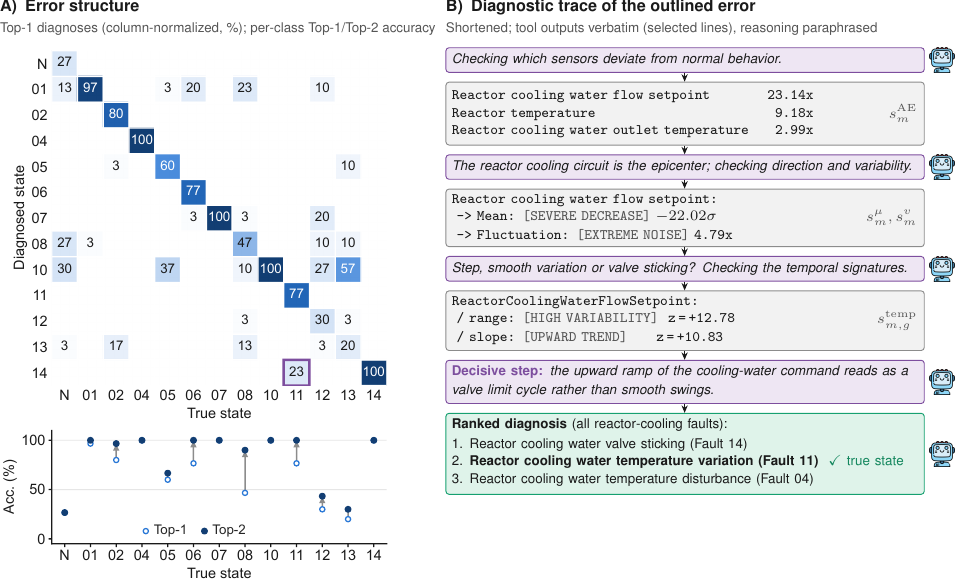}
    \vspace{-0.9cm}
    \caption[Error structure of AgentRCA on TEP]{\textbf{Error structure of AgentRCA on TEP.} \textbf{(A)} Top: Top-1 confusion matrix over three runs, normalized per true state (\%); N denotes normal operation. Bottom: Top-1 (open) and Top-2 (filled) accuracy per true state. Errors concentrate on normal operation and on faults involving random variations or slow drifts (Faults 8, 12, 13). \textbf{(B)} Shortened diagnostic trace for the outlined cell in (A), with tool outputs verbatim (selected lines) and reasoning paraphrased. The agent correctly localizes the anomaly in the reactor cooling circuit but interprets an upward trend of the cooling water flow setpoint as valve sticking (Fault 14). It ranks the true fault (Fault 11, random variation of the cooling water inlet temperature) second, among three reactor-cooling candidates.}
    \label{fig:tep_trace}
    \vspace{-0.4cm}
\end{figure}

Fig.~\ref{fig:tep_trace}A shows that AgentRCA diagnoses step-type faults (Faults 1, 4, 7), Fault 10, and valve sticking (Fault 14) almost perfectly, while errors concentrate on faults involving random variations or slow drifts (Faults 8, 12, 13). These errors are also consistent across runs, with 72\% of misdiagnoses occurring on windows incorrectly classified in every run. Normal operation is particularly difficult to identify, with only 27\% Top-1 accuracy, indicating a tendency to predict faults even for normal windows. One possible explanation is the imbalance in the diagnosis-specific prompt, which devotes substantially more information to fault mechanisms than to normal operation. Top-2 accuracy recovers many errors, particularly for Fault 8 (47\% to 90\%) and Fault 11 (77\% to 100\%), but provides more limited improvement for Faults 12 and 13 and none for normal operation.

Fig.~\ref{fig:tep_trace}B shows a representative error. For a Fault 11 window, the agent identifies abnormal behavior in the reactor cooling circuit but attributes an upward trend in the cooling water flow setpoint to valve sticking (Fault 14), rather than a response to varying inlet temperature. Although the true fault is ranked second, the trace makes the competing physical interpretations and the evidence behind the incorrect ranking examinable. This example illustrates the difficulty of distinguishing fault mechanisms that produce similar responses under closed-loop control.

\subsection{Quantitative evidence and operating-regime matching drive diagnostic accuracy} \label{subsec:results_faithfulness}

\begin{figure}[t]
    \centering
    \vspace{-0.4cm}
    \includegraphics[width=1.0\linewidth]{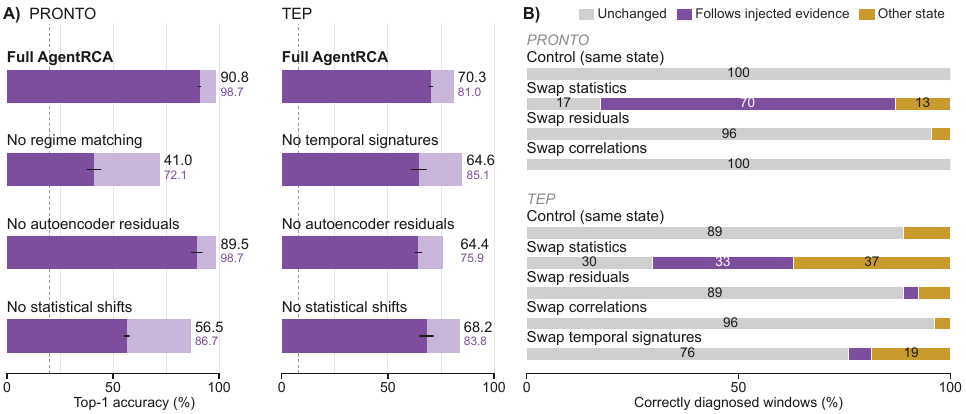}
    \vspace{-0.8cm}
    \caption[Dependence of the diagnosis on the evidence]{\textbf{Dependence of AgentRCA's diagnosis on the evidence returned by its tools.} \textbf{(A)} Top-1 accuracy (dark) and Top-2 accuracy (light) when one evidence source is removed; values give Top-1 (black) and Top-2 (purple) accuracy. Error bars show one standard deviation across runs; dashed lines mark random guessing. \textbf{(B)} Outcome of the evidence swap for windows diagnosed correctly in all three runs (23 on PRONTO, 54 on TEP): the output of one tool is replaced by that tool's output for a window of a different fault (swap) or of the same state (control). Bars show the share of windows whose diagnosis is unchanged, changes to the fault whose evidence was injected, or changes to another state.}
    \label{fig:evidence}
    \vspace{-0.3cm}
\end{figure}

An interpretable diagnostic trace should reflect evidence that influences the agent's decision. We test whether AgentRCA's diagnoses depend on the supplied quantitative evidence through two complementary experiments (Fig.~\ref{fig:evidence}): removing individual evidence sources and replacing their outputs with evidence from other windows.

\paragraph{Evidence ablation}
We remove one evidence source at a time and repeat the evaluation with DeepSeek-V4.1-Flash (Fig.~\ref{fig:evidence}A). On both datasets, we remove either the autoencoder residuals or the statistical shifts. On PRONTO, we additionally remove operating-regime matching, so the statistical and correlation tools use a global normal reference pooled over all setpoints. Because TEP uses a single normal reference, we instead remove its temporal-signature tool. The removed tool is also omitted from the recommended tool order; the rest of the prompt remains unchanged.

On PRONTO, removing regime matching  reduces Top-1 accuracy by 49.8 percentage points (from 90.8\% to 41.0\%), while removing statistical shifts reduces it by 34.3 percentage points to 56.5\%. Removing the autoencoder residuals has little effect (89.5\%). Without regime matching, normal differences between operating setpoints can be mistaken for fault-induced deviations. The resulting drop in accuracy shows that reliable diagnosis requires quantitative evidence evaluated against the operating-condition reference. On TEP, removing individual evidence sources produces smaller reductions in Top-1 accuracy, to 64.6\% without temporal signatures, 64.4\% without autoencoder residuals, and 68.2\% without statistical shifts. Top-2 accuracy remains between 75.9\% and 85.1\%. These smaller reductions are consistent with overlapping diagnostic information across tools, although ablation alone cannot establish how the agent combines their outputs. Reconstruction residuals can identify abnormal multivariate patterns, but appear to contribute less additional accuracy when statistical and temporal deviations already characterize the fault.

\paragraph{Evidence swap}
To test whether quantitative evidence drives otherwise stable diagnoses, we select windows correctly diagnosed in all three runs (up to six per state; 23 on PRONTO and 54 on TEP). For each window, we replace the output of one tool with the output of the same tool for a window representing a different fault. A shift toward the donor fault would indicate that the agent relies strongly on the replaced evidence source. We then assess whether the diagnosis remains unchanged, switches to the fault represented by the replacement evidence, or changes to a third state (Fig.~\ref{fig:evidence}B). All other tool outputs remain unchanged, forcing the agent to reconcile the replacement evidence against potentially  conflicting observations. On PRONTO, donor windows come from the same operating regime; slugging is excluded because no other fault occurs at its setpoints.

As a control, replacing statistical shifts with those from another window of the same state preserves 100\% of PRONTO diagnoses and 89\% of TEP diagnoses. Replacing them with shifts from a different fault changes 83\% of PRONTO diagnoses and 70\% of TEP diagnoses, respectively. On PRONTO, 70\% switch directly to the donor fault, compared with 33\% on TEP, where another 37\% shift to a third state, reflecting conflicts with unchanged tool outputs. Swapping autoencoder residuals or correlation changes leaves 89–100\% of diagnoses unchanged, while swapping temporal signatures changes 24\% on TEP. 

These results show that AgentRCA responds to fault-specific quantitative evidence, with directional statistical shifts having the strongest influence on the final diagnosis. The smaller effects of replacing residuals and correlations suggest that these outputs rarely overturn the evidence from other tools, rather than lacking diagnostic information. The relative influence of each tool varies across systems and fault mechanisms.

\subsection{Language model and reasoning protocol} \label{subsec:results_ablations}

\begin{figure}[t]
    \centering
    \includegraphics[width=1.0\linewidth]{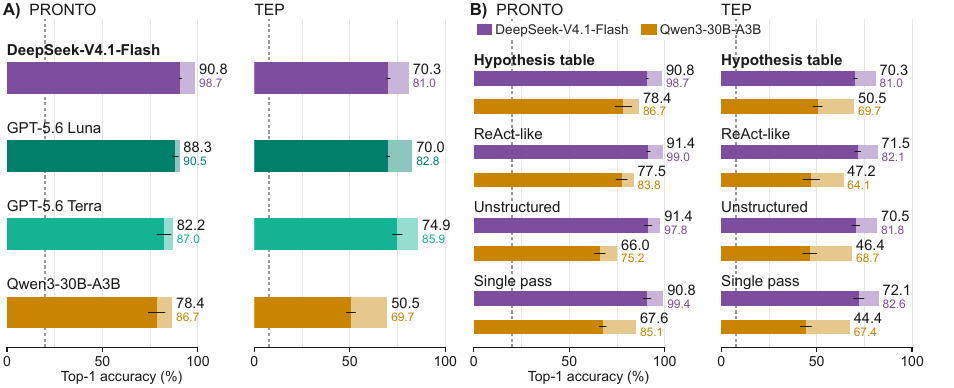}
    \vspace{-0.8cm}
    \caption[Language model and reasoning protocol]{\textbf{Effect of the language model and the reasoning protocol on accuracy.} \textbf{(A)} Top-1 (dark) and Top-2 (light) accuracy for each LLM with the hypothesis-table protocol. \textbf{(B)} Accuracy for each reasoning protocol with DeepSeek-V4.1-Flash (purple) and Qwen3-30B-A3B-Thinking-2507 (amber). Values give Top-1 (black) and Top-2 (colored) accuracy. Error bars show one standard deviation across runs; dashed lines mark random guessing.}
    \label{fig:models_protocols}
\end{figure}

\paragraph{Language model} \label{subsec:model_robustness}
To assess AgentRCA's dependence on the underlying LLM, we evaluate three additional models using identical tools and prompts (Fig.~\ref{fig:models_protocols}A): GPT-5.6 Luna~\cite{openai2026gpt56}, GPT-5.6 Terra, and the smaller open-weight Qwen3-30B-A3B-Thinking-2507~\cite{qwen3thinking2507}. On TEP, the three larger models achieve comparable Top-1 accuracy ($70.0\%$-$74.9\%$), with Top-2 accuracies between 81.0\% and 85.9\%. On PRONTO, GPT-5.6 Luna performs similarly to DeepSeek-V4.1-Flash in Top-1 accuracy (88.3\% vs.\ 90.8\%) but achieves lower Top-2 accuracy (90.5\% vs.\ 98.7\%), while GPT-5.6 Terra reaches 82.2\% Top-1 and 87.0\% Top-2. The smaller Qwen3-30B-A3B-Thinking-2507 model also substantially outperforms the LLM baseline on both datasets, although its Top-1 accuracy drops to 78.4\% on PRONTO and 50.5\% on TEP. Overall, these results demonstrate that AgentRCA performs well across different LLM backbones, although their relative performance varies between benchmarks. The larger performance gap on TEP suggests that interpreting evidence from more complex processes requires stronger diagnostic reasoning capabilities.

\newpage
\paragraph{Reasoning protocol}
To assess whether AgentRCA's diagnostic performance depends more on reasoning structure or diagnostic evidence, we compare four protocols using the same diagnostic tools and engineering knowledge (Fig.~\ref{fig:models_protocols}B): the default hypothesis-table protocol, a ReAct-like protocol that justifies each tool call without maintaining a hypothesis table, an unstructured protocol where the agent can call tools without restrictions, and a single-pass protocol in which all tool outputs are computed in advance and the LLM answers in one response. With DeepSeek-V4.1-Flash, the four configurations reach similar accuracy on both datasets. On PRONTO, Top-1 accuracy ranges from 90.8\% to 91.4\%, with Top-2 accuracy between 97.8\% and 99.4\%. On TEP, Top-1 accuracy ranges from 70.3\% to 72.1\%, with Top-2 accuracy between 81.0\% and 82.6\%. Notably, both the unstructured protocol, in which the agent queries tools without explicit reasoning guidance, and the single-pass protocol, in which all evidence is provided at once, achieve accuracy comparable to the structured iterative protocols. These results suggest that, with a sufficiently capable LLM, access to informative quantitative evidence can yield high diagnostic accuracy without prescribing how that evidence is acquired or organized.

With the smaller Qwen3-30B-A3B-Thinking-2507, reasoning structure has a stronger effect on PRONTO: the hypothesis-table and ReAct-like protocols achieve 78.4\% and 77.5\% Top-1 accuracy, compared with 66.0\% and 67.6\% for the unstructured and single-pass protocols. On TEP, however, all four protocols achieve between 44.4\% and 50.5\% Top-1 accuracy, suggesting that additional reasoning structure cannot fully compensate for the difficulty of interpreting evidence from a more complex process. We retain the hypothesis-table protocol as the default because it explicitly tracks supporting and contradicting evidence for competing fault mechanisms, producing a structured diagnostic trace that can be inspected after the diagnosis.

\subsection{Engineering knowledge in the prompt and in technical documentation} \label{subsec:results_knowledge}

\begin{figure}[t]
    \centering
    \vspace{-0.8cm}
    \includegraphics[width=1.0\linewidth]{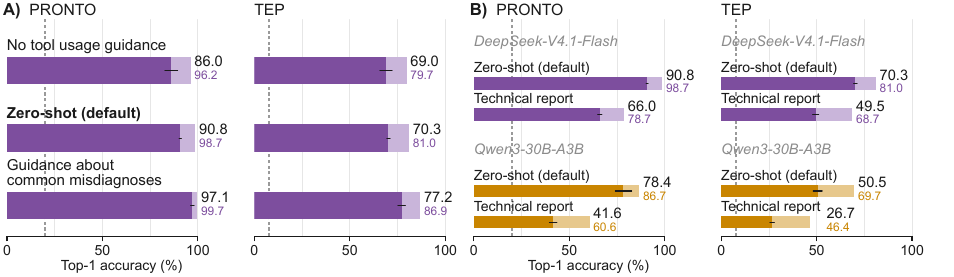}
    \vspace{-0.8cm}
    \caption[Effect of engineering knowledge]{\textbf{Effect of the engineering knowledge available to AgentRCA.} \textbf{(A)} Top-1 (dark) and Top-2 (light) accuracy with DeepSeek-V4.1-Flash for three prompt variants: without guidance on tool usage, with the default zero-shot prompt, and with additional guidance about common misdiagnoses. \textbf{(B)} Accuracy with the default zero-shot prompt and when diagnosis-specific system and fault descriptions are replaced by technical documentation, for DeepSeek-V4.1-Flash (purple) and Qwen3-30B-A3B-Thinking-2507 (amber). Values give Top-1 (black) and Top-2 (colored) accuracy. Error bars show one standard deviation across runs; dashed lines mark random guessing.}
    \label{fig:knowledge}
\end{figure}

\paragraph{Prompt content}
We vary the engineering knowledge provided in the prompt with DeepSeek-V4.1-Flash (Fig.~\ref{fig:knowledge}A). Removing guidance on how to interpret the diagnostic tools, while retaining the system and fault descriptions, has little effect: Top-1 accuracy decreases from 90.8\% to 86.0\% on PRONTO and from 70.3\% to 69.0\% on TEP. Consistent with the reasoning-protocol ablation, this suggests that a capable LLM can exploit quantitative evidence without extensive procedural guidance. Conversely, adding a few domain-informed rules targeting common misdiagnoses increases Top-1 accuracy to 97.1\% on PRONTO and 77.2\% on TEP (99.7\% and 86.9\% Top-2). These rules address, for example, the distinction between air leakage and air blockage on PRONTO and the tendency to attribute unexplained variability to Fault 10 on TEP. Because these rules were developed after inspecting errors on labeled evaluation windows, this experiment moves beyond the strictly zero-shot setting to an error-informed prompt refinement. The improvements illustrate how diagnostic behavior can be adjusted through natural-language rules without retraining the digital twin or fine-tuning the LLM, but do not establish generalization to unseen fault windows.

\paragraph{Technical documentation}
Finally, we assess whether AgentRCA can derive the engineering context needed for diagnosis directly from technical documentation (Fig.~\ref{fig:knowledge}B). We replace the structured system and fault descriptions in its prompt with the complete technical document used by the LLM-without-tools baseline, while retaining the candidate state names, recommended tool order, and execution rules. With DeepSeek-V4.1-Flash, Top-1 accuracy decreases from 90.8\% to 66.0\% on PRONTO and from 70.3\% to 49.5\% on TEP; with Qwen3-30B-A3B-Thinking-2507, it decreases from 78.4\% to 41.6\% and from 50.5\% to 26.7\%. Nevertheless, diagnostic tools retain substantial value even without structured fault descriptions: with the same technical documentation, the tool-augmented DeepSeek-V4.1-Flash outperforms the LLM baseline by 38 percentage points on PRONTO and 16 points on TEP. Organizing engineering knowledge around candidate fault mechanisms provides an additional gain of 25 and 21 points, respectively. These results indicate that access to technical documentation alone cannot replace explicit descriptions linking fault mechanisms to their expected physical signatures. The larger performance drop with Qwen on PRONTO suggests that smaller LLMs may struggle to extract and connect relevant engineering information from unstructured documentation to quantitative evidence.

Overall, accurate zero-shot diagnosis benefits from combining quantitative evidence of deviations from normal operation with engineering knowledge organized around candidate fault mechanisms. Neither raw measurements nor unstructured technical documentation achieves comparable diagnostic performance in the evaluated configurations.

\section{Discussion} \label{sec:discussion}

AgentRCA moves beyond identifying abnormal variables or assigning fault labels by evaluating which physical fault mechanism best explains the observed system behavior. Its central contribution is the integration of tool-based quantitative analysis with engineering-guided hypothesis evaluation, connecting numerical observations to physical explanations. The results on PRONTO and TEP demonstrate that this approach can support accurate diagnosis without labeled fault examples, using normal-operation measurements and qualitative descriptions of fault mechanisms. This is particularly relevant for rare or safety-critical faults, where labeled examples are scarce.\vspace{3pt}

The experiments highlight the complementary roles of quantitative evidence and engineering knowledge. Providing raw measurements to an LLM yields substantially lower accuracy than supplying diagnostic evidence extracted through dedicated tools. Organizing engineering knowledge around candidate faults improves diagnostic accuracy by connecting observed deviations to plausible physical mechanisms. The evidence interventions further show that AgentRCA's decisions depend on quantitative evidence rather than linguistic reasoning alone. The comparable performance of different reasoning protocols suggests that explicit reasoning structure is not essential for accurate diagnosis, at least with sufficiently capable LLMs. Nevertheless, structured protocols allow selective evidence gathering and provide a record of how competing hypotheses were assessed. AgentRCA's diagnostic performance thus relies on the combination of quantitative tools and engineering knowledge, rather than on a particular prompting protocol.\vspace{3pt}

A practical advantage of AgentRCA is its ability to incorporate engineering knowledge expressed in natural language rather than requiring a complete mathematical model or an exhaustive diagnostic rule base. Such knowledge is typically available in industrial facilities through system specifications, operating manuals, technical reports, and failure mode analyses. Our experiments show that organizing this information into diagnostic-specific descriptions is important for effective use by the LLM. This allows maintenance engineers to refine or extend these descriptions directly in natural language, without retraining the digital twin or fine-tuning the LLM.\vspace{3pt}

\newpage
Transparency and traceability are particularly important for AI-based diagnosis in industrial environments. Unlike conventional classifiers that predict fault labels or anomaly-attribution methods that identify abnormal variables, AgentRCA produces diagnoses supported by traceable quantitative evidence. By exposing this evidence and retaining alternative hypotheses, AgentRCA allows engineers to assess and challenge its conclusions. Although the diagnostic trace does not guarantee that generated explanations faithfully reflect the LLM’s internal decision process, it makes supporting and contradicting evidence available for inspection. AgentRCA is therefore intended to support expert decision-making alongside existing monitoring and safety systems.

AgentRCA relies on normal-operation references, predefined candidate fault mechanisms, and accurate interpretation of quantitative evidence through engineering knowledge. Continuously changing operating conditions may complicate the characterization of normal behavior, while unknown or simultaneous faults may fall outside the available hypotheses. Diagnostic accuracy also depends on the quality of the engineering descriptions and the LLM's ability to interpret the available evidence. Future work could improve normal-operation modeling, extend diagnosis to unknown fault mechanisms, and incorporate physics-based consistency checks, additional temporal analyses, or targeted diagnostic tests to distinguish mechanisms with overlapping signatures. Validation across further industrial systems would help assess the broader applicability of evidence-grounded diagnosis.

\section{Conclusion} \label{sec:conclusion}

We proposed AgentRCA, a framework that combines a data-driven digital twin trained on normal-operation measurements with engineering knowledge and tool-grounded LLM hypothesis evaluation for industrial root cause analysis. Experiments on PRONTO and TEP demonstrate its ability to diagnose physical fault mechanisms without historical labeled fault examples, achieving competitive accuracy when fault supervision is scarce. Ablations and evidence interventions show that diagnostic performance depends on quantitative evidence, appropriate operating-condition references, and diagnosis-specific engineering knowledge. AgentRCA connects observed deviations to candidate physical mechanisms through an inspectable diagnostic process, providing engineers with both a proposed cause and the evidence and reasoning supporting it. 


\section*{Ethics and inclusion statement}

AgentRCA is intended to support, not replace, expert judgment in industrial fault diagnosis and root cause analysis. Its outputs should be reviewed by qualified personnel before any operational intervention, particularly in safety-critical settings. Potential risks include erroneous diagnoses, overreliance on generated explanations, and incomplete or inaccurate source documentation. We recommend validation on representative scenarios, evidence traceability, and uncertainty reporting before real-world deployment.

\newpage
\section*{Data availability}

This study did not generate any new materials. The PRONTO heterogeneous benchmark dataset used in this study is publicly available from Zenodo at \url{https://doi.org/10.5281/zenodo.1341583}. The Tennessee Eastman process simulation dataset used in this study is publicly available from Harvard Dataverse at \url{https://doi.org/10.7910/DVN/6C3JR1}.

\section*{Code availability}

The code needed to reproduce the analyses will be released in a public repository upon acceptance of the paper. The source code is available at \url{https://github.com/AmauryWEI/agentrca}.

\section*{Acknowledgements}
The authors gratefully acknowledge Prof. Derek Nowrouzezahrai for hosting A.W. during a visiting research stay at MILA Quebec Artificial Intelligence Institute (Montreal, Canada), where this research was conducted.

\section*{Funding}
This work was financially supported by the Swiss National Science Foundation (SNSF) Grant Number 200021\_200461.

\section*{CRediT authorship contribution statement}
A.W. and O.F. conceptualized the idea, A.W. developed the methodology, A.W. and O.F. conceived the experiments, A.W. conducted the experiments, A.W. and O.F. analyzed the results, and A.W. and O.F. wrote the manuscript.

\section*{Declaration of competing interest}
The authors declare no competing interests.

\bibliographystyle{elsarticle-num}
\bibliography{manual}

\end{document}